\documentclass[letterpaper, 10 pt, conference]{ieeeconf}  
\usepackage{cite}
\IEEEoverridecommandlockouts                              % This command is only needed if 
\usepackage{amsmath,amsfonts,bm}
\usepackage{float}

\usepackage[table]{xcolor}
\usepackage{times}
\usepackage{epsfig}
\usepackage{graphicx}
\usepackage{amssymb}
\usepackage{url}            % simple URL typesetting
\usepackage{booktabs}       % professional-quality tables
\usepackage{nicefrac}       % compact symbols for 1/2, etc.
\usepackage{multirow}
\usepackage{bigstrut}
\usepackage{graphicx}
\usepackage{amsmath}
\usepackage{amssymb}
\usepackage{algorithm}
\usepackage{algorithmicx}
\usepackage{algpseudocode}
\usepackage{tabu}
\usepackage{siunitx}
\usepackage{adjustbox}
\usepackage{siunitx}
\usepackage{makecell}
\usepackage{colortbl}
\usepackage{color}
\usepackage{multirow}
\usepackage{blindtext}
\usepackage{comment}
\usepackage{bm}
\usepackage{bbm}
\usepackage{booktabs}
\usepackage{wrapfig}
\usepackage{array} % for extended column definitions, e.g. >{\em}c
\newcommand{\qsq}{$\mathrm{VQ}^{2}\!\mathrm{A}$\xspace}

\usepackage{wrapfig}

\title{\LARGE \bf
FIQ: Fundamental Question Generation with the Integration of Question Embeddings for Video Question Answering}

\author{Ju-Young Oh$^{1}$, Ho-Joong Kim$^{1}$, and Seong-Whan Lee$^{1}$% 
\thanks{*This research was supported by the Institute of Information \& Communications Technology Planning \& Evaluation (IITP) grant, funded by the Korea government (MSIT) (No. RS-2019- II190079 (Artificial Intelligence Graduate School Program (Korea University)), and No. RS-2024-00457882 (AI Research Hub Project)).}
\thanks{
$^{1}$J.-Y. Oh, H.-J. Kim, and S.-W. Lee are with the Department of Artificial Intelligence, Korea University, Anam-dong, Seongbuk-ku, Seoul 02841, Korea.
\texttt{\small \{juyoungoh, hojoong\_kim, sw.lee\}@korea.ac.kr}
}}

\begin{document}

\maketitle
\thispagestyle{empty}
\pagestyle{empty}

%%%%%%%%%%%%%%%%%%%%%%%%%%%%%%%%%%%%%%%%%%%%%%%%%%%%%%%%%%%%%%%%%%%%%%%%%%%%%%%%
\begin{abstract}

Video question answering (VQA) is a multimodal task that requires the interpretation of a video to answer a given question. Existing VQA methods primarily utilize question and answer (Q\&A) pairs to learn the spatio-temporal characteristics of video content. However, these annotations are typically event-centric, which is not enough to capture the broader context of each video. The absence of essential details such as object types, spatial layouts, and descriptive attributes restricts the model to learning only a fragmented scene representation. This issue limits the model's capacity for generalization and higher-level reasoning. In this paper, we propose a fundamental question generation with the integration of question embeddings for video question answering (FIQ), a novel approach designed to strengthen the reasoning ability of the model by enhancing the fundamental understanding of videos. FIQ generates Q\&A pairs based on descriptions extracted from videos, enriching the training data with fundamental scene information. Generated Q\&A pairs enable the model to understand the primary context, leading to enhanced generalizability and reasoning ability. Furthermore, we incorporate a VQ-CAlign module that assists task-specific question embeddings with visual features, ensuring that essential domain-specific details are preserved to increase the adaptability of downstream tasks. Experiments on SUTD-TrafficQA demonstrate that our FIQ achieves state-of-the-art performance compared to existing baseline methods.


\end{abstract}

%%%%%%%%%%%%%%%%%%%%%%%%%%%%%%%%%%%%%%%%%%%%%%%%%%%%%%%%%%%%%%%%%%%%%%%%%%%%%%%%
\section{INTRODUCTION}
% \cite{BSN} 
Video question answering (VQA) is a multimodal task~\cite{kim2024te} that combines computer vision and natural language processing. It requires the model to answer given questions based on the understanding of dynamic events in a video. VQA has acquired significant attention for its importance of tasks and broad applications in various fields, including education, health care, and surveillance systems~\cite{lee2015motion}. Despite the substantial advancements of existing works in recent years and the wide applicability of the task, the alignment of natural languages and visual features still remains as a challenge. Recent studies have demonstrated significant advancements in this area, with various works ~\cite{videocontextalignertransformer, blip2, lee2024text} achieving notable results by aligning both modalities.

\begin{figure}[!t]

\includegraphics[width=0.95\linewidth]{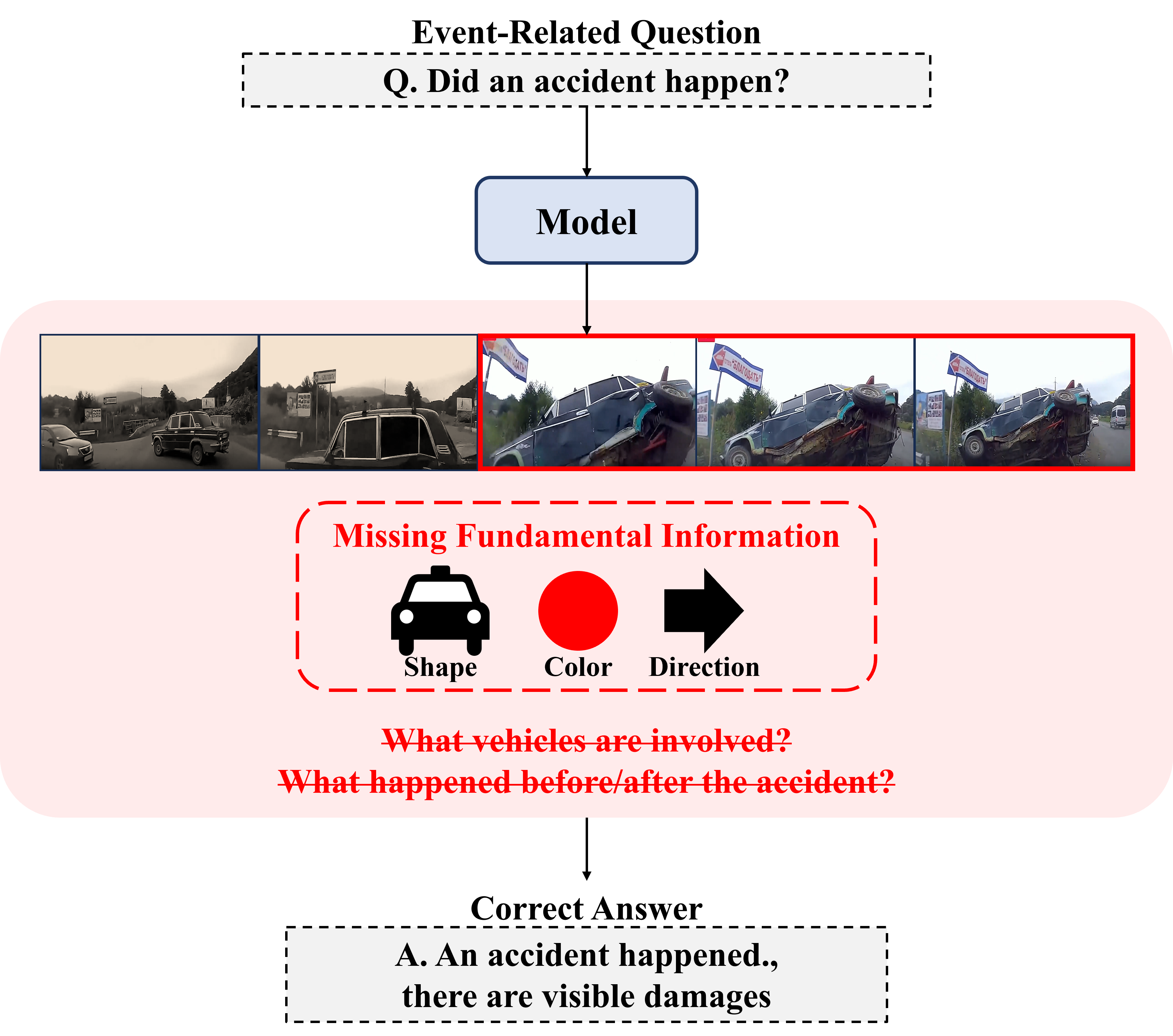}

\caption{The existing dataset only focuses on event-centric information of video, but not on fundamental information of video such as shape, color, and direction of objects.}
\label{fig:issue}
% \end{center}
\vspace{-0.4cm}
\end{figure}

The existing VQA methods employ CLIP-based encoders to leverage their image-text alignment capability from their pretrained knowledge of large-scale data. Even though video-based encoders~\cite{videollava, VATT, slowfast, videomae} which are specialized for video data exist, visual-text alignment requires both pretrained visual and text encoders, and CLIP provides both of them. FrozenBiLM~\cite{frozenBiLM} introduces a lightweight module that connects the frozen image encoder from CLIP and the frozen bidirectional language model for effective multimodal reasoning through masked language modeling. ViLA~\cite{vila} proposes QFormer-Distiller, a module to enhance the alignment of both modalities by teaching Q-Former from BLIP~\cite{blip}. While CLIP offers strong cross-modal features, it is pretrained on static images and thus relies heavily on textual annotations to supply spatio-temporal context. However, the current VQA dataset usually provides event-centric textual annotations, frequently omitting the fundamental scene attributes such as object identity, shape, or color. Although event-centric annotations already provide semantic cues, they serve only partial scene representations, thereby limiting a model to acquire only a fragmentary understanding of each scene.

Fig.~\ref{fig:issue} shows an example in which a VQA model exclusively trained on event-centric annotations easily focuses on partial scenes. The model focuses on partial scenes where the collision occurs while preceding and subsequent frames are largely ignored. This behavior occurs because the event-centric annotations alone provide insufficient information to answer the corresponding questions. Such reliance on event-centric scenes restricts the ability of the model to both generalization and higher-level reasoning, which requires a broader understanding of the video context~\cite{lee1997new, lee1999integrated}. 

As shown in Fig.~\ref{fig:issue}, the model recognizes that an accident occurs, but fails to identify which vehicles are involved or what events happened before or after the accident. This partial understanding of video from the model is not enough for the model to establish causal or temporal relations between frames, leading to weaker generalization and limited higher-level reasoning. To alleviate this issue, enriching the fundamental attributes such as shape, color, and the direction of objects is significant in enhancing the reasoning ability of the model. These attributes are crucial for enabling the model to track other attributes over time, thereby facilitating a broader understanding of the video content, rather than limiting the comprehension to specific event-related frames.

We propose a fundamental question generation with the integration of question embeddings for video question answering (FIQ), which integrates the general Q\&A pairs into the original dataset to enhance the fundamental video understanding. Additionally, we incorporate question embeddings as task-specific information through a VQ-CAlign module. Our approach integrates Q\&A pairs that focus on the fundamental understanding of video such as object types and shapes in specific scenes to the original dataset, enabling the model to enhance the fundamental understanding of video for more advanced reasoning. Furthermore, we introduce the VQ-CAlign module to integrate the task-specific information using question embeddings. This module prevents the model from missing the necessary task-specific information to interpret the direction of the task after the integration of general information. 

Our contributions are summarized as follows:
\begin{itemize}
    \item We propose FIQ which generates Q\&A pairs to provide a fundamental understanding of the visual information, enhancing the generalizability of the model.
    \item We construct the VQ-CAlign module to provide question embeddings to reflect task-specific information.
    \item Our method achieves the state-of-the-art result on SUTD-TrafficQA dataset compared to other baseline models.
\end{itemize}

\section{RELATED WORKS}

\subsection{Video Question Answering}
VQA is the task of determining the semantic information of a video to generate an answer to a given query. There are two approaches designed to achieve a lower computational cost for the VQA task, adapter-based methods, and utilizing the text-based representations. Adapter-based methods~\cite{clip-adapter, vita, graphadapter} are proposed to reduce the computational cost of VQA tasks, thereby enabling large language models (LLM) to be adapted to downstream tasks without the need of finetuning the entire model. Tem-Adapter~\cite{temAdapter} presents a novel alignment method that utilizes auto-regression to integrate semantic and temporal information from the video domain to the image domain. While various works adapt textual information to enhance the understanding of spatial and temporal features from video, achieving a competitive understanding of video data solely through textual representations for VQA is not common. Vamos~\cite{wang2024vamos} is a text-based video understanding framework that achieves superior performance without relying on visual features by generating task-agnostic texts, emphasizing utilizing only textual data leads to enhanced performance. Similarly, ColPro~\cite{cai2024empowering} integrates three distinct task-specific prompts to prevent catastrophic forgetting during the training process. Despite the differences in objective, both methods demonstrate remarkable performance without relying on visual embeddings. Our approach aligns with these two previous methods by leveraging textual prompts to enhance the interpretability of video data, reflecting its spatial and temporal properties. Furthermore, we employ both a language model (LM) and a LLM to generate task-agnostic Q\&A pair to provide an overview of the fundamental components of the video.

\subsection{General Question Generation}
Question generation (QG) is a task aimed at automatically generating questions, aims to expand semantic diversity and extract insights beyond visually explicit content~\cite{lim2000text, lee1990translation}. There are two directions in QG, generating task-specific questions and general questions. QG tasks for providing task-specific information demonstrate state-of-the-art results in various methods since the question is one of the most task-specific inputs to provide guidance to a model about how to interpret the data. Prophet~\cite{pllm} and SGSH~\cite{sgsh} propose knowledge base question generation (KBQG) methods which generates natural language questions using external knowledge beyond the given images. While QG is helpful in assisting with task-specific information, it could be used as an effective data augmentation method to provide more general questions such as colors, objects, and their number. \qsq~\cite{changpinyo2022all} supports robust multilingual capability and integrates multiple models to produce diverse Q\&A pairs, by primarily using the T5 model. Additionally, an all-in-one QAG model~\cite{ushio2023practical} highlights the potential of textual captions to enrich visual question answering datasets by incorporating details not explicitly depicted in visual representations. Building on these approaches, we employ LMs to generate contextually rich and temporally informed questions. We utilize both LM and LLM to show a more adaptable approach.

\begin{figure*}[t!]
\begin{center}
% \centerline{\includegraphics[width=\linewidth]{method_figure.png}}
\includegraphics[width=\linewidth]{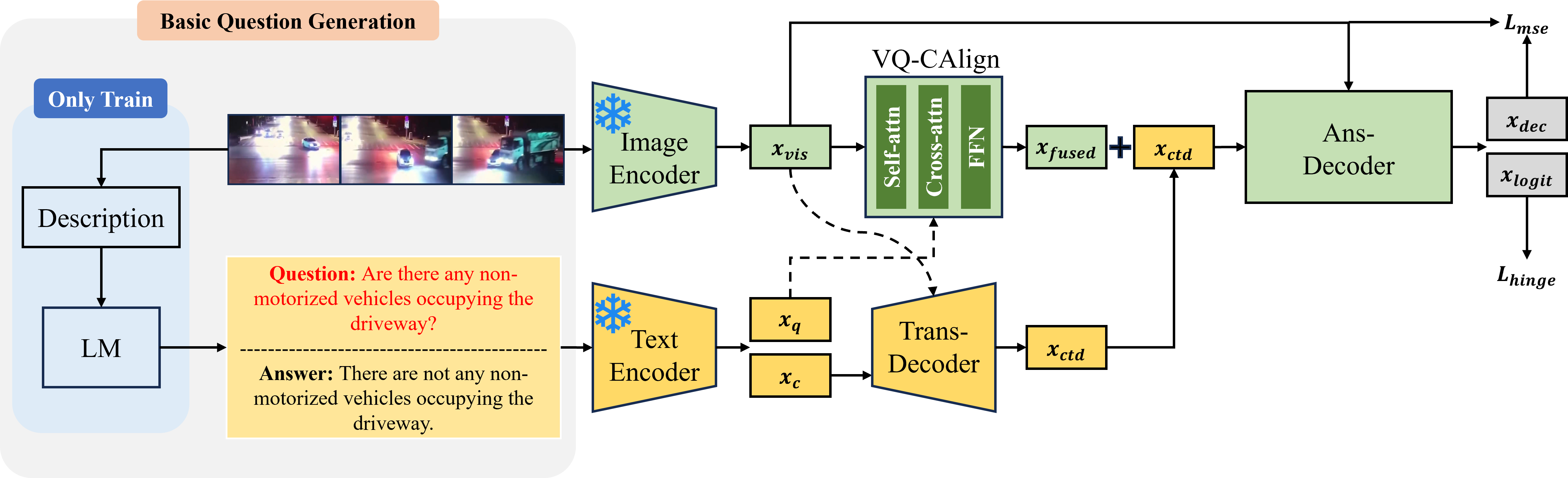}
% \includegraphics[width=\linewidth]{fig_test.pdf}

%\vspace{-0.3cm}
\caption{Overall architecture of FIQ. It consists of four pivotal sub-processes. Q\&A pair which contains the general information of video first generated using language model such as T5~\cite{t5}, and GPT~\cite{gpt4}. The frozen text encoder takes these generate Q\&A pairs with the original dataset as an input, and each question embeddings and answer candidate embeddings are passed to the Trans-Decoder and VQ-CAlign. The frozen image encoder takes video data as input, and extracted visual features are passed to VQ-CAlign with question embeddings. Both modalities are merged, and passed to the Ans-Decoder, which fuses visual and textual information to align the temporal information.}

\label{fig:main}
\end{center}
%\vspace{-0.8cm}
\end{figure*}
\section{METHOD}

FIQ consists of four main processes: Fundamental question generation, textual representation refinement, integration of question embeddings, and visual representation alignment. Fig.~\ref{fig:main} shows the overall process of FIQ. The subsequent subsections will provide a detailed explanation of each process.

\subsection{Preliminaries} 
The objective of the multi-choice VQA task is to identify the best answer $a_{final}$ from the options presented in the given question $x_{q}$ and the visual feature $x_{vis}$. The score for each answer candidate $x_{c}$ is calculated, and the answer with the highest matching score will be the final answer of the VQA task. The final predicted answer $\hat{a}_{final}$ is derived as follows:

\begin{equation} \label{eq:vqa}
% \hat{\rva} = \argmax_{\rva \in \mathcal{A}} \mathcal{F} (\rva|\rvv,\rvq),
\hat{a}_{final} = \text{argmax}(x_{c}|x_{vis},x_{q}).
% _{final}
\end{equation} 

\subsection{Fundamental Question Generation}\label{qnaGenmethod}

We employ VideoChat2~\cite{mvbench} for generating descriptions that cover both low-level features such as color and objects and high-level features such as motions or temporal orders from video. For extracted descriptions, we filter the repetitive same numbers that are not relevant to given video. From filtered descriptions, we employ LMs to generate Q\&A pairs, such as T5 and GPT-4o-mini. Following \qsq~\cite{changpinyo2022all}, we provide a prompt to follow three steps: candidate answer extraction, question generation, and answer validation. 

\subsubsection{Candidate Answer Extraction}
For the candidate answer extraction, we guide LM to extract candidate answers, including noun phrases, named entities, short open-class word sequences, boolean literals yes/no, and object counts including zero when no count is mentioned. 

\subsubsection{Question Generation}

The LM generates questions by rewriting the source sentence into an interrogative form for candidate answers. To encourage a diversity of questions, we instruct LM to cover various types of questions such as ``How many..", ``Where are..", and ``Is there.." based on the given candidate answers. All questions and answers are generated under 77 tokens to ensure the maximum input length accepted by the pretrained CLIP text encoder.

\begin{table*}[!t]
\caption{
Performance comparison with state-of-the-art methods on SUTD-TrafficQA and each (H) and (H$^*$) represent training prompts with and without adapter heads. (H) and (A) represent methods for adding prompts, respectively. Avg represents an average accuracy for all six tasks.}

\centering
\begin{tabular}{l| c c c c c c c}
% \hline
% \hline
\toprule
\multirow{2.5}{*}{\textbf{Methods}} &\multicolumn{7}{c}{\textbf{SUTD-TrafficQA}} \\ 
\cmidrule(lr){2-8}
~ &\multicolumn{1}{c}{\textbf{B}} & \multicolumn{1}{c}{\textbf{F}} &  \multicolumn{1}{c}{\textbf{R}} &  \multicolumn{1}{c}{\textbf{C}} & 
\multicolumn{1}{c}{\textbf{I}} &  \multicolumn{1}{c}{\textbf{A}} &
\multicolumn{1}{c}{\textbf{Avg}}
%  & & & & & & & & & & & &
\\
\midrule
Unsupervised CLIP~\cite{clip} & 25.6 & 20.1 &34.0   & 30.8 & 22.8 & 28.8 &  26.5   \\ 
CLIP~\cite{clip} + Template & 31.8 & 36.0 & 29.9 & 71.8   & 22.1 &  33.4 & 32.3   \\ 
Totally finetuning & 39.8 & 35.1 & 46.6 & 45.6 & 37.2  & 40.5  & 40.3   \\ 
Partially finetuning  & 41.6 & 37.8 &  44.6  & 50.0  & 33.1 & 41.7 & 41.7 \\ 
LoRA~\cite{lora} & 38.7 & 38.7 &  36.7  & 37.9  & 34.5 & 38.1 & 38.3  \\
CLIP-Adapter~\cite{clip-adapter} &  35.8 &  32.0 &  35.4 &  42.3 & 33.1   &32.1  &34.8    \\ 
Multi-layer Adapter~\cite{clip-adapter} &  30.5 &  26.6& 26.5 & 38.5 & 28.3  & 25.8 & 29.1   \\ 
% ATP~\cite{atp} &  &  &  & &  &  &35.6 & 93.2 \\ 
Prompt learning (H)~\cite{coop} & 42.4  &  32.4  &   45.2  & 55.5  &  40.7 & 43.6   & 42.9   \\ 
Prompt learning (H$^*$)~\cite{coop} &  40.3 & 33.2   &  41.0   &  46.5  & 34.9  &   38.4 &39.7  \\
Prompt learning (A)~\cite{vpt} &  41.7  &  31.5 &  40.1 & 48.4  & 33.1  &  41.4 & 41.1 \\ 
Tem-Adapter~\cite{temAdapter} & 45.5 & 37.2 &45.8   & 54.5 & 35.1 & 48.3 &  46.0 \\
\midrule
FIQ &  \textbf{46.9} &   \textbf{43.5} &  \textbf{52.5} &    \textbf{54.0} &   \textbf{39.8}  &   \textbf{51.8} &  \textbf{48.4}    \\ 
\bottomrule
% \hline
\end{tabular}
% \end{center}
\label{table:base}
%\vspace{-0.8cm}
\end{table*}

\subsubsection{Answer Validation}
To validate generated Q\&A pairs, we utilize a token-level F1 score~\cite{tokenlevel} to validate whether the candidate's answer matches the original sentence. When the score is lower than 0.54, we discard that sample to ensure its correctness. These generated Q\&A pairs have a one-to-one pair for each question and one single answer. Since the SUTD-TrafficQA dataset requires a multi-choice format, every question should provide multiple options as an answer to ensure seamless integration with the original dataset. We designed the positive answer to be an answer to the question from the target video ID, and the rest of the negative answers to be sampled from different video IDs. To preserve the randomness of the negative answers, we randomly select three distinct video IDs and randomly pick one answer from their multiple available answers to serve as a negative option. We integrate these generated Q\&A pairs into the original dataset for training.
\subsection{FIQ}\label{visproc}
\subsubsection{Textual Representation Refinement} \label{textproc} 

We employ a frozen text encoder and a Trans-Decoder to focus on extracting meaningful information from the textual input. We use a frozen CLIP~\cite{clip} text encoder to independently extract textual embeddings from both the questions and the answer candidates consisting of four candidate options. In parallel, we obtain visual representations from a frozen image encoder, denoted as $x_{vis} \in \mathbb{R}^{N \times D}$, where $N$ is the number of video frames and $D$ is the feature dimension of the CLIP image encoder. We also extract answer candidate embeddings, denoted as $x_{c} \in \mathbb{R}^{T \times D}$, where $T$ is the sequence length of the textual data. These two inputs, $x_{vis}$ and $x_{c}$, are then passed into the Trans-Decoder to produce refined answer candidate embeddings, denoted as $x_{ctd}\in \mathbb{R}^{T \times D}$.

\subsubsection{Integration of Question Embeddings}\label{qalign}

We apply learnable positional embeddings to capture the dynamic information of video. Learnable positional embeddings are expressed as follows:

\begin{equation} \label{eq:learnable}
x_{vpe} = x_{vis} + e_{pos}, 
\end{equation}

\noindent where $e_{pos}\in \mathbb{R}^{N \times D} $ is the learnable positional embedding, and $x_{vpe} \in \mathbb{R}^{N \times D}$ is the visual feature that contains the learnable positional information.

Even though general Q\&A pairs provide a fundamental understanding of the video content, still providing task-specific information is important to perform the downstream tasks. We introduce the VQ-CAlign module to fuse the question embeddings with visual embeddings. VQ-CAlign consists of three main modules: self-attention, cross-attention, and feedforward network.  VQ-CAlign takes $x_{vis}$ and question embeddings $x_{q}\in \mathbb{R}^{T \times D}$ as inputs. VQ-CAlign is expressed as follows:

\begin{equation} \label{eq:qatten}
x_{fused} = \text{VQ-CAlign}(x_{vpe}, x_{q}).
\end{equation} 

\noindent Inside the VQ-CAlign, the self-attention takes $x_{vpe}$ as an input and utilizes it as a query, key, and value. It captures the internal relationships and contextual information within the $x_{vpe}$, and generates processed visual embeddings $x_{self}\in \mathbb{R}^{N \times D}$ as an output. Next, $x_{self}$ and $x_{q}$ are passed to the cross-attention module. For cross-attention module, $x_{self}$ serves as a query and $x_{q}$ serves as a key and value. The cross-attention connects $x_{vis}$ and $x_{q}$, allowing the model to focus on visual positions that are relevant to question embeddings, and generates $x_{ca}\in \mathbb{R}^{N \times D}$ as an output. Finally, $x_{ca}$ passes the feedforward network as a final stage, and $x_{fused} \in \mathbb{R}^{N \times D}$ is generated as a final output. 

$x_{fused}$  is added again with $x_{ctd}$ to fully reflect the textual information for the VQA task, integrating the task-specific information better. This process is expressed as follows:

\begin{equation} \label{eq:mix}
x_{mix} = x_{fused} + x_{ctd},
\end{equation} 

\noindent where $x_{mix}\in \mathbb{R}^{N \times D}$ is the fused output of $x_{fused}$ and $x_{ctd}$.

\subsubsection{Visual Representation Alignment}\label{visrepalign}
Sequentially, we adapt the Ans-Decoder to generate the future state by leveraging the historical information. Formally, it is described as:

\begin{equation} \label{eq:transformer}
x_{dec} = \text{Ans-Decoder}(x_{mix}, x_{vis}, x_{vis}), 
\end{equation}

\noindent  where $x_{dec} \in \mathbb{R}^{N \times D}$ is the processed output from Ans-Decoder. Finally, $x_{dec}$ is used to calculate the cosine similarity by taking ${x_{dec}}^{\top}$ and $x_{ctd}$ as inputs, and we obtain $x_{logit}\in \mathbb{R}^{N}$ as an output of the answer with the best matching score to the question.

\subsection{Training and Inference}

\subsubsection{Training}

For the final loss, we adapt the addition of Hinge loss and MSE loss. The overall loss function is described as follows:

\begin{equation} \label{eq:Finalloss}
L_{final} = \gamma L_{hinge}(x^{-}_{logit}, x^{+}_{logit}) +  L_{mse}(x_{dec}, x_{vis})\text{,}
\end{equation}

\noindent  where $\gamma$ is a ratio to balance both loss functions and $x^{+}_{logit}$ is the ground truth answer of ${{x}^{-}_{logit}}$. Hinge loss takes $x_{logit}$ and ${{x}^{-}_{logit}}$ as inputs and MSE loss takes $x_{dec}$ and $x_{vis}$ as inputs.

\subsubsection{Inference}
In the inference, we only use the original dataset without generating the additional dataset. We first extract both visual and textual embeddings and align them in visual and textual respects as described in Sec.~\ref{textproc}, Sec.~\ref{qalign} and Sec.~\ref{visrepalign}. Lastly, we calculate the $x_{logit}$ for the final prediction.

\section{EXPERIMENTS}
\subsection{Setup}

\begin{table*}[!t]
\caption{Ablation studies on the SUTD-TrafficQA by adding the VQ-CAlign and the dataset generated by T5 and GPT. Avg represents an average accuracy for all six tasks.}

\centering
\begin{tabular}{l| c c c c c c c}
% \hline
\toprule
\multirow{2}{*}{\textbf{Methods}} &\multicolumn{7}{c}{\textbf{SUTD-TrafficQA}}\\ 
\cmidrule(lr){2-8}
~ &\multicolumn{1}{c}{\textbf{B}} & \multicolumn{1}{c}{\textbf{F}} &  \multicolumn{1}{c}{\textbf{R}} &  \multicolumn{1}{c}{\textbf{C}} & 
\multicolumn{1}{c}{\textbf{I}} &  \multicolumn{1}{c}{\textbf{A}} 
&\multicolumn{1}{c}{\textbf{Avg}}
%  & & & & & & & & & & & &
\\
\midrule
Tem-Adapter~\cite{temAdapter} & 45.5 & 37.2 &45.8   & 54.5 & 35.1 & 48.3 &  46.0  \\ 
VQ-CAlign & 44.8 & 46.1 & 47.1 & 51.3 & 33.7  & 50.1  & 46.3  \\ 
\midrule
VQ-CAlign + T5~\cite{t5}  &   \textbf{46.1} &  \textbf{47.0} &   \textbf{52.1} &   \textbf{58.3}  &   \textbf{35.8} &  \textbf{50.9} & \textbf{47.8}  \\ 
VQ-CAlign + GPT~\cite{gpt4} &  \textbf{46.9} &   \textbf{43.5} &  \textbf{52.5} &   \textbf{54.0} &   \textbf{39.8}  &   \textbf{51.8} &  \textbf{48.4} \\ 
\bottomrule
% \hline
\end{tabular}
% \end{center}
\label{table:ablation}
\end{table*}

\subsubsection{Hyperparameters}
During preprocessing, we use CLIP~\cite{clip} with a ViT/B-16 as the backbone and set the visual feature dimension as 512. We extract 8 clips of 16 consecutive frames from each video, a total of 128 frames. For training, we set the batch size as 32 and the epoch as 37. Additionally, we set the exponential moving average rate to 0.9999. We apply a cosine decay schedule with a rate of 2. We apply a learnable embedding with a dropout rate of 0.2 and a maximum sequence length of 128. For attention modules, we set the number of heads to 16.

\subsubsection{Dataset} 
We conduct experiments on SUTD-TrafficQA dataset. The dataset consists of 10,080 videos and annotated 62,535 QA pairs annotated by humans, respectively. SUTD-TrafficQA focuses on traffic scenarios, requiring to understand of the specific traffic events and the causal relations related to them. It allows the evaluation of event understanding ability and causal inference for cognition of the model. SUTD-TrafficQA consists of six types of different reasoning tasks related to traffic scenarios. 

\noindent\textbf{Basic Understanding (B).}
This task requires the model to interpret the basic level of traffic scenarios, including feature-query, event-query, event classification, and counting.

\noindent\textbf{Event Forecasting (F).}
This task evaluates the forecasting ability of the models based on the outcome of the current scene. The question for forecasting is given to a model, and the model reasons future events for a given video.

\noindent\textbf{Reverse Reasoning (R).}
This task involves finding the events that happened prior to the beginning of a video segment.

\noindent\textbf{Introspection (I).}
This task tests the model to check whether the model enables it to provide preventive advice related to the accident that occurs in the given video. 

\noindent\textbf{Attribution (A).}
This task evaluates the model's ability to reason about the causes of traffic events by selecting the underlying factors from given answer candidates.

\noindent\textbf{Counterfactual Inference (C).}
This task differs in its objective from previously introduced tasks, as it requires reasoning about hypothetical scenarios not given in the video. The task requires the model to reason about imagery events based on the given conditions in the question.

\subsection{Main Results} 
Our goal is to generate Q\&A pairs that contain fundamental information from the video to enhance the reasoning ability of the model. Even though SUTD-TrafficQA already contains Q\&A pairs related to basic information, we add more of it by using LM. This integration of Q\&A pairs shows overall improvement of the performance. Table~\ref{table:base} shows the comparison with state-of-the-art methods. Compared to other state-of-the-art methods, our method achieves outstanding performance on SUTD-TrafficQA. Our method shows overall improvement on five given tasks. Especially, it shows notable improvement in F, R, I, and A tasks. All four tasks require the understanding of events that actually happened in the video, which requires a factual inference to answer a question within the given information in the video. 

These results show that our generated Q\&A pairs serve the necessary fundamental information for reasoning tasks F, R, I, and A. Even though Q\&A pairs related to basic attributes already exist in current data, existing data was not enough to strengthen the reasoning ability to perform tasks such as F, R, I, and A. These results show that although we prompt LM to generate fundamental questions, the resulting Q\&A pairs inherently contain spatio-temporal information. This stems from the extracted descriptions reflecting the spatio-temporal context of the video as discussed in Section.~\ref{qnaGenmethod}. From these results, extracted video descriptions enable the generated Q\&A pairs to supply the fundamental information of temporal dynamics that are crucial for reasoning tasks.

Compared to other tasks, C remains almost the same accuracy because of its distinguished direction compared to other tasks. C requires hypothetical reasoning beyond the given context of the video rather than the understanding of the given spatio-temporal information of the video. Due to this inherent difference in reasoning requirements, task C shows a different accuracy trend.

\begin{figure}[!t]
% \begin{center}
% \centerline{\includegraphics[width=0.95\linewidth]{method_figure_v2.png}}
\centering
\includegraphics[width=0.8\linewidth]{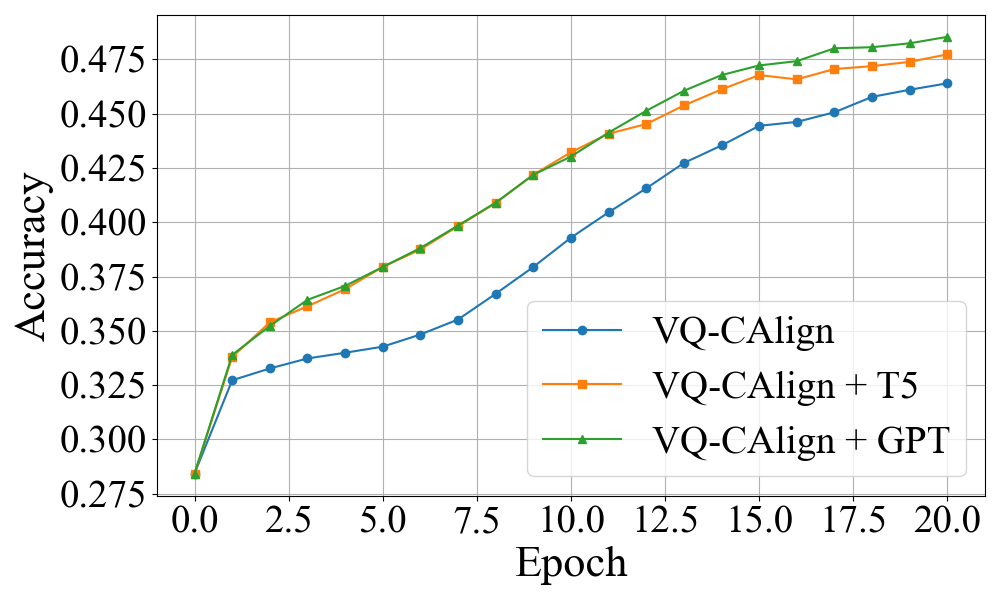}
%\vspace{-0.3cm}
\caption{Comparison between different LM-based Q\&A generation (T5, GPT) methods on SUTD-TrafficQA.}

\label{fig:accGraph}
% \end{center}
\vspace{-0.4cm}
\end{figure}

\subsection{Ablation Studies} 
To demonstrate the effectiveness of our model, we conduct an ablation study on each key component. Table~\ref{table:ablation} shows the performance improvements by adding each module. We introduce a VQ-CAlign module to integrate question embeddings as task-specific features. Compared to the Tem-adapter~\cite{temAdapter} that is our baseline model, we observe a meaningful improvement in the accuracy by utilizing the VQ-CAlign module. Beyond architectural modifications, we incorporate an additional Q\&A pair that contains the fundamental information to the dataset. To generate these fundamental Q\&A pairs, we employed both T5 and GPT for comparative analysis. Although generated Q\&A pairs generated from T5 shows an improved performance, its practical application in real-life scenarios is limited due to the availability of pretrained LLMs. We observe that GPT-generated Q\&A pairs achieve 48.4\% accuracy, which is the best performance compared to T5-generated Q\&A pairs. It highlights LLM's effectiveness in capturing the primary attributes of video data, compared to LM. Fig.~\ref{fig:accGraph} illustrates the overall accuracy improvements achieved by adding each module and the generated Q\&A pairs. As shown in Fig.~\ref{fig:accGraph}, all three settings of FIQ converge at epoch 20 which shows the fast convergence speed.

\section{CONCLUSION}
In this paper, we propose FIQ, a framework that enhances video reasoning by introducing fundamental Q\&A pair generation method and VQ-CAlign mechanisms. We generate fundamental Q\&A pairs for assisting event-centric textual annotations by leveraging LMs to improve the model's reasoning ability and generalizability. Additionally, the VQ-CAlign module integrates task-specific knowledge through question embeddings to better support downstream VQA tasks. Experiment results show that our approach significantly improves the accuracy for the reasoning-related tasks, which proves the integration of general knowledge of video enhances the reasoning ability of the model compared to other existing methods. In future work, we plan to generate a new dataset that reflects the question information inside the dataset as answer candidates.

\bibliographystyle{IEEEtran}
\bibliography{REFERENCE}

\end{document}